\documentclass[11pt]{article}
\usepackage{geometry}
\usepackage{coling2020}
\usepackage{times}
\usepackage{url}
\usepackage{amssymb}
\usepackage{amsmath}
\usepackage{latexsym}
\usepackage{microtype}
\usepackage{graphicx}
\usepackage{enumitem}
\usepackage[dvipsnames]{xcolor}
\definecolor{example_blue}{HTML}{007FFF}

\usepackage{booktabs}
\colingfinalcopy 

\title{BUT-FIT at SemEval-2020 Task 5:  Automatic detection of counterfactual statements with deep pre-trained language representation models}

\author{Martin Fajcik, Josef Jon, Martin Docekal, Pavel Smrz\\
  Brno University of Technology, Faculty of Information Technology\\
  612\,66 Brno, Czech Republic \\
  {\tt \{ifajcik,ijon,idocekal,smrz\}@fit.vutbr.cz} }
\date{}

\begin{document}
\maketitle
\begin{abstract}
  This paper describes BUT-FIT's submission at SemEval-2020 Task 5: Modelling Causal Reasoning in Language: Detecting Counterfactuals. The challenge focused on detecting whether a given statement contains a counterfactual (Subtask 1) and extracting both antecedent and consequent parts of the counterfactual from the text (Subtask 2). We experimented with various state-of-the-art language representation models (LRMs).  We found RoBERTa LRM to perform the best in both subtasks. We achieved the first place in both exact match and F1 for Subtask 2 and ranked second for Subtask 1.
\end{abstract}

\section{Introduction}
\blfootnote{
    %
    %
    %
    %
    \hspace{-0.65cm}  
    This work is licensed under a Creative Commons 
    Attribution 4.0 International Licence.
    Licence details:
    \url{http://creativecommons.org/licenses/by/4.0/}.
    %
    %
}

\begin{figure}[ht!]
    \centering
    \includegraphics[width=\columnwidth]{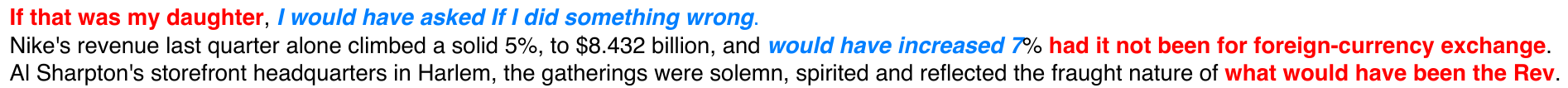}
    \caption{Three examples from the training data containing counterfactual statements. Antecedents are highlighted with red bold, consequents with blue bold italic. The last example has no consequent.}
    \label{fig:cntf_example}
\end{figure}

One of the concerns of SemEval-2020 Task 5: Modelling Causal Reasoning in Language: Detecting Counterfactuals \cite{yang2020} is to research the extent to which current state-of-the-art systems can detect counterfactual statements. 
A counterfactual statement, as defined in this competition, is a conditional composed of two parts. 
The former part is the antecedent -- a statement that is contradictory to known facts.
The latter is the consequent -- a statement that describes what would happen had the antecedent held\footnote{
According to several definitions in literature, e.g.  \cite{starr2019}, the antecedent of counterfactual might not need to counter the facts.
}.
To detect a counterfactual statement, the system often needs to posses a commonsense world knowledge to detect whether the antecedent contradicts with it. 
In addition, such a system must have an ability to reason over consequences that would arise had the antecedent would have been true. 
In some cases, the consequent might not be present at all, but instead a sequence resembling consequent, but with no consequential statement, might be present. Figure \ref{fig:cntf_example} shows a set of examples drawn from the data.

Counterfactuals were studied across a wide spectrum of domains. For instance, logicians and philosophers focus on logical rules between parts of counterfactual and its outcome \cite{goodman1947}. Political scientists conducted counterfactual thought experiments for hypothetical tests on historical events, policies or other aspects of society \cite{tetlock1996}. However, there is only a small amount of work in computational linguistics studying this phenomenon. SemEval-2020 Task 5 aims at filling this gap in the field. The challenge consists of two subtasks:
\begin{enumerate}
\item Detecting counterfactual statements -- classify whether the sentence has a counterfactual statement.
\item Detecting antecedent and consequence -- extract boundaries of antecedent and consequent from the input text.
\end{enumerate}

The approaches we adopted follow recent advancements from deep pre-trained language representation models. In particular, we experimented with fine-tuning of BERT \cite{devlin2019}, RoBERTa \cite{liu2019} and ALBERT \cite{lan2019} models. 
Our implementation is available online\footnote{\url{https://github.com/MFajcik/SemEval_2020_Task-5}}.

\section{System overview}
\subsection{Language Representation Models}
We experimented with three language representation models (LRMs):

\textbf{BERT} \cite{devlin2019}
 is pre-trained using the multi-task objective consisting of denoising LM and inter-sentence coherence (ISC) sub-objectives. The LM objective aims at predicting the identity of 15\% randomly masked tokens present at the input\footnote{The explanation of token masking is simplified and we refer readers to read details in the original paper \cite{devlin2019}.}. Given two sentences from the corpus, the ISC objective is to classify whether the second sentence follows the first sentence in the corpus. The sentence is replaced randomly in half of the cases.
During the pre-training, the input consists of two documents, each represented as a sequence of tokens divided by special $[SEP]$ token and preceded by $[CLS]$ token used by the ISC objective, i.e. $[CLS]document_1[SEP]document_2[SEP]$. The input tokens are represented via jointly learned token embeddings ${E_t}$, segment embeddings $E_s$ capturing whether the word belongs into $document_1$ or $document_2$ and positional embeddings ${E_p}$ since self-attention is position-invariant operation. During fine-tuning, we leave the second segment empty.

\textbf{RoBERTa} \cite{liu2019} is a BERT-like model with the different training procedure. This includes dropping the ISC sub-objective, tokenizing via byte pair encoding \cite{sennrich2016} instead of WordPiece, full-length training sequences, more training data, larger batch size, dynamic token masking instead of token masking done during preprocessing and more hyperparameter tuning.

\textbf{ALBERT} \cite{lan2019} 
 is a RoBERTa-like model, but with n-gram token masking (consecutive n-grams of random length from the input are masked), cross-layer parameter sharing, novel ISC objective that aims at detecting whether the order of two consecutive sentences matches the data, input embedding factorization, SentencePiece tokenization \cite{kudo2018} and much larger model dimension. The model is currently at the top of leaderboards for many natural language understanding tasks including GLUE \cite{wang2018} and SQuAD2.0 \cite{rajpurkar2018}.

\subsection{Subtask 1: Detecting counterfactual statements}
The first part of the challenge is a binary classification task, where the participating systems determine whether the input sentence is a counterfactual statement.

A baseline system applying an SVM classifier \cite{cortes1995} over TF-IDF features was supplied by the organizers.  We modified this script to use other simple classifiers over the same features -- namely Gaussian Naive Bayes and 6-layer perceptron network, with 64 neurons in each layer.

As a more serious attempt at tackling the task, we compare these baselines with state-of-the-art LRMs -- RoBERTa and ALBERT. The input is encoded the same way as in \ref{systemo:subtask2}.  We trained both models with cross-entropy objective and we used the linear transformation of \texttt{CLS}-level output after applying dropout for classification.  After the hyperparameter search, we found that RoBERTa model performed the best on this task. For our final system, we built an ensemble from the best checkpoints of RoBERTa model.

\subsection{Subtask 2: Detecting antecedent and consequence}
\label{systemo:subtask2}
We extended each LRM in the same way \newcite{devlin2019} extended BERT for SQuAD. The input representation for input $x$ is obtained by summing the input embedding matrices $\boldsymbol{E} = \boldsymbol{E}_t + \boldsymbol{E}_s + \boldsymbol{E}_p \in \mathbb{R}^{L \times d_i}$ representing its word embeddings $\boldsymbol{E}_t$, position embeddings $\boldsymbol{E}_p$ and segment embeddings\footnote{RoBERTa is not using segment embeddings.} $\boldsymbol{E}_s$ with $L$ being the input length and $d_i$ input dimensionality. Applying LRM and dropout $\delta$, an output matrix $\delta(LRM(E)) = \boldsymbol{H} \in \mathbb{R}^{L \times d_o}$ is obtained, $d_o$ being the LRM's output dimensionality. Finally, a linear transformation is applied to obtain logit vector for antecedent start/end $a_s$, $a_e$ and consequent start/end $c_s$, $c_e$.
\begin{equation} \label{eq:model2}
P_j(\cdot|x) \propto \exp{(\boldsymbol{w}_j^{\top}\boldsymbol{H})}; j \in \{a_{s}, a_{e}, c_{s}, c_{e}\}
\end{equation}
For consequent, we do not mask \texttt{CLS}-level output and use it as a no consequent option for both $c_{s}$ and $c_{e}$. Therefore we predict that there is no consequent iff model's prediction is $c_{s}=0$ and $c_{e}=0$; assuming $0$ is the index of \texttt{CLS}-level output. Finally, the log-softmax is applied and model is trained via minimizing cross-entropy for each tuple of inputs $x$ and target indices $t$ from the dataset $\mathcal{D}$.
\begin{equation} \label{eq:xe_task2}
- \sum_{ (x,t) \in \mathcal{D}} \sum_{j \in \{a_{s}, a_{e}, c_{s}, c_{e}\}} \log P_j(t_j|x)
\end{equation}
An ensemble was built using a greedy heuristic seeking the smallest subset from the pool of trained models s.t. it obtains best exact match on a validation set
\footnote{For more details on how the ensemble was built,  see \texttt{TOP-N} fusion in \newcite{fajcik2019}.}.

\section{Experimental setup}
\subsection{Data}
For each subtask, training datasets without split were provided. Therefore we took the first 3000 examples from Subtask 1 data and 355 random examples from Subtask 2 data as the validation data. The train/validation/test split was 10000/3000/7000 for Subtask 1 and 3196/355/1950 for Subtask 2. 88.2\% of the validation examples in Subtask 1 were labeled 0 (non-counterfactual).
\subsection{Preprocessing \& Tools}
In case of Subtask 1, after performing a length analysis on the data, we truncated input sequences at length of 100 tokens for the LM based models in order to reduce worst-case memory requirements, since only 0.41\% of the training sentences were longer than this limit. A histogram of the example lengths in tokens is presented in Appendix \ref{app:data_analysis}. For Subtask 2, all the input sequences fit the maximum input length of 509 tokens. 

 For the preliminary experiments with simpler machine learning methods, we adopted the baseline script provided by the organizers, which is based on  \texttt{sklearn} Python module. We implemented our neural network models in \textit{PyTorch} \cite{paszke2019} using \textit{transformers} \cite{wolf2019} library. In particular, we experimented with \textit{roberta-large} and \textit{albert-xxlarge-v2} in Subtask 1 and with \textit{bert-base-uncased}, \textit{bert-large-uncased}, \textit{roberta-large} and \textit{albert-xxlarge-v1} models in Subtask 2. We used \textit{hyperopt} \cite{bergstra2013} to tune model hyperparameters. See Appendix \ref{app:hyperparameters} for further details on hyperparameters. We used the Adam optimizer with a decoupled
weight decay \cite{loshchilov2017}. For Subtask 2, we combined this optimizer with lookahead \cite{zhang2019}. All models were trained on 12GB GPU.

\section{Results and analysis}
For \textbf{Subtask 1}, we adapted the baseline provided by the task organizer to asses how more classical machine learning approaches perform on the dataset. After seeing the subpar performance, we turned our attention to pre-trained LRMs, namely RoBERTa and ALBERT. The results of the best run of each model can be found in Table \ref{tab:task1_results}. A more comprehensive list of results for different hyperparameters can be found in the Appendix \ref{app:subt1_hyperp}.

Our final submission is an ensemble of RoBERTa-large models since we found that this LRM performs better than ALBERT for this task.  We trained a number of models on the train set and computed F1 scores on the validation part. 10 best (in terms of F1) single models were selected, and the output probabilities were averaged for all the possible combinations of these models. The combination with highest F1 score was selected as a final ensemble. Then we trained new models with the same parameters as the models in the ensemble, but using the whole training data, including the part that was previously used for validation. Finally, for our submitted ensemble, we used checkpoints saved after the same number of updates as the best checkpoints for the systems trained only on part of the training data.

\begin{table}[t!]
\begin{center}

\begin{tabular}{|l|ccc|}
\hline \bf Model & \bf Precision & \bf Recall & \bf F1 \\ \hline
SVM & 80.55 & 8.19 & 14.87 \\ \hline
Naive Bayes & 22.81 &28.81 & 25.47 \\ \hline
MLP & 39.01 & 29.09 & 33.33 \\ \hline

RoBERTa-large & $\boldsymbol{98.11} \pm0.11$ & $\boldsymbol{85.71} \pm0.61$ & $\boldsymbol{91.48}\pm0.51$ \\ \hline
ALBERT-xxlarge & $97.99  \pm0.15$ & $84.19 \pm0.55$   & $90.59  \pm0.67$\\ \hline
RoBERTa-large-ens & $98.95$ & 87.30 & $92.76$\\ \hline

\end{tabular}
\end{center}
\caption{
\label{tab:task1_results} Results of different models on validation part of Subtask 1 training data (first 3000 sentences). Results for RoBERTa and ALBERT models are averaged over ten training runs with the best found hyperparameters.}
\end{table}

We performed an error analysis of the best single RoBERTa and ALBERT models. RoBERTa model misclassified 52 examples (29 false positives, 23 false negatives), while ALBERT misclassified 60 examples (32 false positives, 23 false negatives). 29 wrongly classified examples were common for both of the models. Examples of wrongly classified statements are presented in the Appendix \ref{app:wrongly_classified}.

\begin{table}[ht!]

\begin{center}
\begin{tabular}{|c|c c c c|c|}
\hline
 & BERT-base    & BERT-large  & ALBERT               & RoBERTa                  & RoBERTa$_{(Ensemble)}$ \\ \hline
\textbf{\textbf{EM}}         & $73.03\pm0.4$ & $72.68\pm0.6$ & $73.59\pm0.8$          & $\boldsymbol{73.92}\pm0.3$ & $74.93$                \\ \hline
\textbf{\textbf{F1}}         & $87.13\pm0.4$ & $87.31\pm0.5$ & $88.13\pm0.4$          & $\boldsymbol{88.27}\pm0.4$ & $80.80$                \\ \hline
\textbf{A}$_{EM}$            & $77.18$     & $76.56$     & $\boldsymbol{77.97}$ & $77.58$                  & $79.44$                \\ \hline
\textbf{A}$_{F1}$            & $91.11$     & $90.93$     & $\boldsymbol{91.71}$ & $91.33$                  & $91.96$                \\ \hline
\textbf{C}$_{EM}$            & $68.87$     & $68.70$     & $69.21$              & $\boldsymbol{70.28}$     & $70.42$                \\ \hline
\textbf{C}$_{F1}$            & $83.15$     & $83.70$     & $84.54$              & $\boldsymbol{85.23}$     & $85.65$                \\ \hline
\textbf{ACC}$_{no-c}$        & $90.37$     & $91.10$     & $91.49$              & $\boldsymbol{91.69}$     & $91.83$                \\ \hline
$\boldsymbol{\#\theta}$      & $109$M      & $335$M      & $223$M               & $355$M                   & \multicolumn{1}{c|}{-} \\ \hline
\end{tabular}
\end{center}
\caption{Results on the Subtask 2 validation data. For EM/F1, we report means and standard deviations. The statistics were collected from 10 runs. $\#\theta$ denotes the number of model's parameters. We also measured EM/F1 for the extraction of antecedent/consequent separately; denoted as A$_{EM}$, A$_{F1}$ and C$_{EM}$, C$_{F1}$ respectively. At last ACC$_{no-c}$ denotes no-consequent classification accuracy.}
\label{tab:task2_results}
\end{table}

For \textbf{Subtask 2}, the results are presented in Table \ref{tab:task2_results}. The hyperparameters were the same for all LRMs.  An ensemble was composed of 11 models drawn from the pool of 60 trained models. We found the ALBERT results to have a high variance. In fact, we recorded our overall best result on validation data with ALBERT, obtaining $75.35$/$89.00$ EM/F1. However, in competition, we submitted only RoBERTa models due to less variance and slightly better results on average\footnote{We submitted the best ALBERT model in the post-evaluation challenge phase, obtaining worse test data results than the ensemble.}.

\section{Related work}
Closest to our work, \newcite{son2017} created a counterfactual tweet dataset and built a pipeline classifier to detect counterfactuals. The authors identified 7 distinct categories of counterfactuals and firstly attempted to classify the examples into one of these categories using a set of rules. Then for certain categories, they used a linear SVM classifier \cite{cortes1995} to filter out tricky false positives.

A large effort in computational linguistics was devoted to the specific form of counterfactuals -- so-called \textit{what-if} questions.  A recent paper by \newcite{Tandon2019WIQAAD} presents a new dataset for what-if question answering, including a strong, BERT-based baseline. The task is to choose an answer to a hypothetical question about cause and an effect, e.g. \textit{Do more wildfires result in more erosion by the ocean?}. Each question is accompanied by a paragraph focused on the topic of the question, which may or may not contain enough information to choose the correct option. The authors show that there is still a large performance gap between humans and state-of-the-art models (73.8\% accuracy for BERT against 96.3\% for a human). This gap is caused mainly by the inability of the BERT model to answer more complicated questions based on indirect effects, which require more reasoning steps. However, the results show that the BERT model was able to answer a large portion of the questions even without accompanying paragraphs, indicating that the LRM models have a notion of commonsense knowledge. 

\section{Conclusions}
We examined the performance of current state-of-the-art language representation models on both subtasks and we found yet another NLP task benefits from unsupervised pre-training. In both cases, we found RoBERTa model to perform slightly better than other LRMs, while its results also being more stable. We have ended up first in both EM and F1 on Subtask 2 and second in Subtask 1.

\section*{Acknowledgements}
This work was supported by the Czech Ministry of Education, Youth and Sports, subprogram INTERCOST, project code: LTC18006.

\bibliographystyle{coling}
\bibliography{semeval2020}

\appendix

\section{Supplemental Material}
\label{sec:supplemental}
\subsection{Hyperparameters}

\label{app:hyperparameters}
\subsubsection{Subtask 1}
The results of RoBERTa models with their training hyperparameters are presented in Table \ref{app:subt1_hyperp}.
\begin{table}[!ht]
\centering
\begin{tabular}{c|c|c|c}
\textbf{batch size} & \textbf{learning rate}     & \textbf{best acc }    &  \textbf{best F1}      \\ \toprule
48         &  2.00E-05 &  0.9829943314 &  0.9209302326 \\
70         &  1.00E-05 &  0.9823274425 &  0.9180834621 \\
72         &  2.00E-05 &  0.9829943314 &  0.9199372057 \\
90         &  4.00E-05 &  0.9829943314 &  0.9209302326 \\
90         &  1.00E-05 &  0.9783261087 &  0.8992248062 \\
96         &  3.00E-05 & \textbf{0.9839946649} & \textbf{0.9240506329} \\
120        &  3.00E-05 &  0.9809936646 &  0.9107981221 \\
132        &  4.00E-05 &  0.9796598866 &  0.9060092450 \\

\end{tabular}
\caption{Different batch sizes and learning rates used to train RoBERTa-large models, results of the best checkpoint on the validation part of the data.}
\label{app:subt1_hyperp}
\end{table}

\noindent We kept other RoBERTa model hyperparameters as shown in Table \ref{app:subt1_hyperp2} for all training runs.
\begin{table}[ht!]
\centering
\begin{tabular}{|c|c|}
\hline
\textbf{Hyperparameter}         & \textbf{Value} \\ \hline
Max gradient norm       &    1.0     \\ \hline
    Epochs         &         8 \\ \hline
     Maximum input length                & 100         \\ \hline
    Dropout & 0.1  \\ \hline
    Optimizer & Adam ($\epsilon$ = 1e-8) \\ \hline
\end{tabular}
\caption{Hyperparameters for Subtask 2, shared for all runs. }
\label{app:subt1_hyperp2}
\end{table}

\subsubsection{Subtask 2}
Our tuned hyperparameters are in Table \ref{tab:hyp_t2}. All other hyperparameters were left the same as PyTorch's default. We did not use any learning rate scheduler.
\begin{table}[ht!]
\centering
\begin{tabular}{|c|c|}
\hline
\textbf{Hyperparameter}         & \textbf{Value} \\ \hline
Dropout rate (last layer)       & 0.0415         \\ \hline
Lookahead K                     & 1.263e-5       \\ \hline
Lookahead $\alpha$ & 0.470          \\ \hline
Max gradient norm               & 7.739          \\ \hline
Batch size                      & 64             \\ \hline
Weight Decay                    & 0.02           \\ \hline
Patience                        & 5              \\ \hline
Max antecedent length           & 116            \\ \hline
Max consequent length           & 56             \\ \hline
\end{tabular}
\caption{Hyperparameters for Subtask 2. We tune only dropout at the last layer (the dropout mentioned in \ref{systemo:subtask2}). Patience denotes the maximum number of epochs, after which we stop the training if there was no EM improvement. All parameters were tuned using HyperOpt \cite{bergstra2013}.}
\label{tab:hyp_t2}
\end{table}

\subsection{Data analysis}
\label{app:data_analysis}
\begin{figure}[ht!]
    \centering
    \includegraphics[width=0.75\textwidth, angle=0]{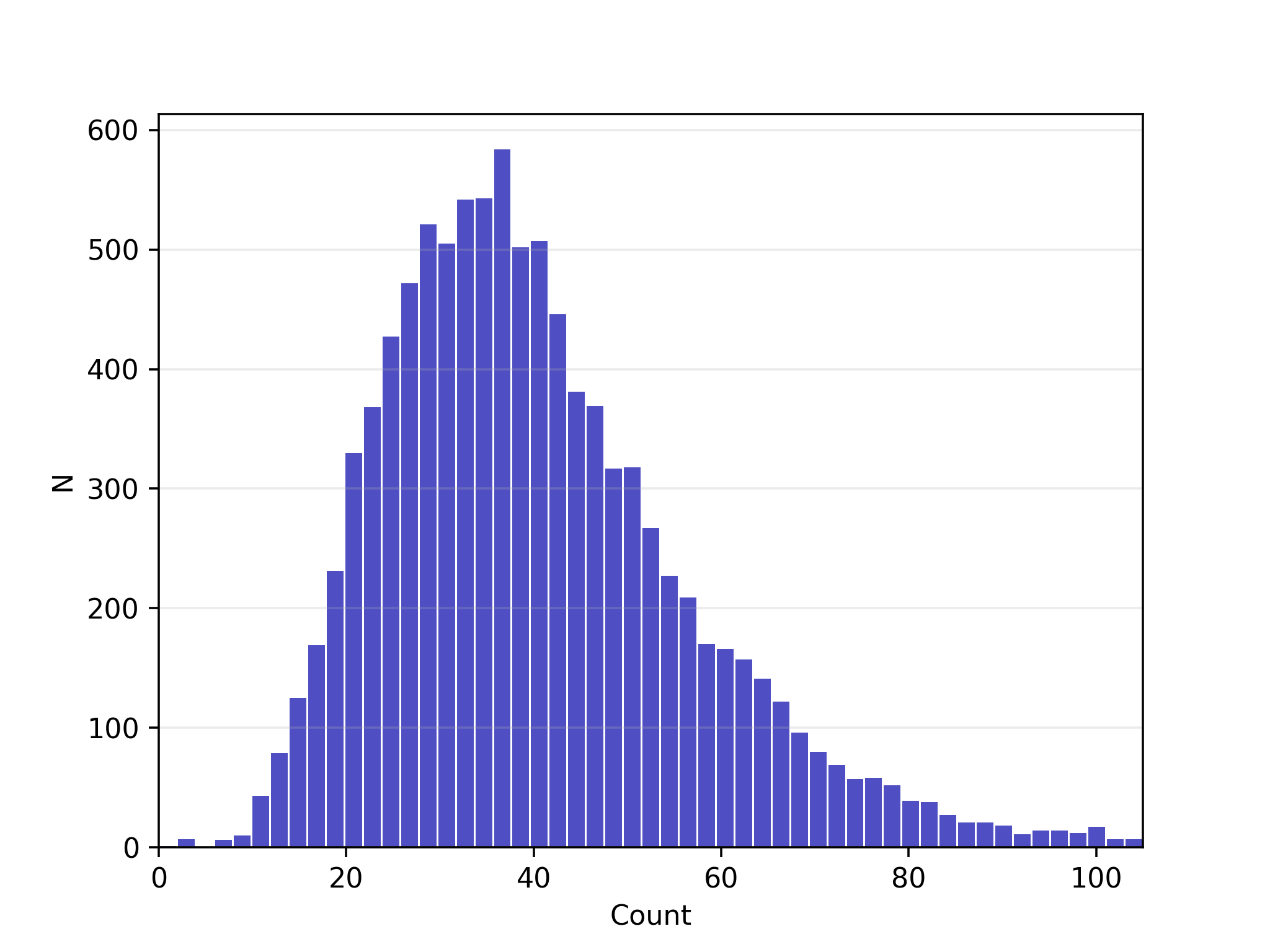}
    \caption{Histogram of example lengths in tokens in the training data for Subtask 1.}
    \label{fig:tok_len}
\end{figure}

The distribution of lengths for examples from Subtask 1 is presented in Figure \ref{fig:tok_len}. We truncate sequences in this subtask to maximum of 100 tokens per example.

\subsection{Wrongly classified examples}
\label{app:wrongly_classified}
Table \ref{app:tab_wrongly_classified_examples} shows examples of statements classified wrongly by both ALBERT and RoBERTa models.

\begin{table}[ht!]
\centering
\begin{tabular}{p{0.75\linewidth}|c|c}
    \textbf{Statement} & \textbf{Predicted}  & \textbf{Correct} \\ \hline
     MAUREEN DOWD VISITS SECRETARY NAPOLITANO - "New Year's Resolutions: If only we could put America in Tupperware": "Janet Napolitano and I hadn't planned to spend New Year's Eve together. & 0  & 1 \\ & & \\
If the current process fails, however, in hindsight some will say that it might have made more sense to outsource the whole effort to a commercial vendor. & 1 & 0 \\
\end{tabular}
\caption{Examples of wrong predictions.}
\label{app:tab_wrongly_classified_examples}
\end{table}

\subsection{Ambiguous labels}

During the error analysis, we noticed a number of examples where we were not sure whether the labels are correct (see Table \ref{app:tab_ambiguous_examples}). 

\begin{table}[ht!]
\begin{center}
\begin{tabular}{p{0.9\linewidth}|c}
    \textbf{Statement}  & \textbf{Label} \\ \hline
     Given that relatively few people have serious, undiagnosed arrhythmias with no symptoms (if people did, we would be screening for this more often), this isn't the major concern. & 0 \\ & \\
     A flu shot will not always prevent you from getting flu, but most will have a less severe course of flu than if they hadn't had the shot," Dr. Morens said. & 0 \\
\end{tabular}
\end{center}
\caption{Examples of ambiguous annotation.}
\label{app:tab_ambiguous_examples}
\end{table}

\subsection{Measurement of results}
The individual measurements for Subtask 2 statistics presented in \ref{tab:task2_results} can be found at \url{https://tinyurl.com/y8zncw7p}. Note that we did not use the same evaluation script as used in official baseline. Our evaluation script was SQuAD1.1 like, ground truth and extracted strings were firstly normalized the same way as in SQuAD1.1, then the strings were compared. For details see our implementation of method \texttt{evaluate\_semeval2020\_task5} in \texttt{scripts/common/evaluate.py}.

\subsection{Wrong predictions in Subtask 2}
\begin{table}[!ht]
\centering
\scalebox{0.87}{
    \begin{tabular}{|p{8cm}|p{8cm}|}
    \hline
    \textbf{Ground Truth}                                                                                                                                                                                                                                                                                                                                                                                     & \textbf{Prediction}                                                                                                                                                                                                                                                                                                                                                                                       \\ \hline
    GLOBAL FOOTPRINT Mylan said in a separate statement that the combination would create "a vertically and horizontally integrated generics and specialty pharmaceuticals leader with a diversified revenue base and a global footprint." On a pro forma basis, \textcolor{red}{\textbf{the combined company would have had revenues of about \$4.2 billion}} and a gross profit, or EBITDA, of about \$1.0 billion in 2006, Mylan said. & GLOBAL FOOTPRINT Mylan said in a separate statement that the combination would create "a vertically and horizontally integrated generics and specialty pharmaceuticals leader with a diversified revenue base and a global footprint." \textcolor{red}{\textbf{On a pro forma basis}}, \textcolor{example_blue}{\textbf{\textit{the combined company would have had revenues of about \$4.2 billion and a gross profit, or EBITDA, of about \$1.0 billion in 2006}}}, Mylan said. \\ \hline
    \textcolor{example_blue}{\textbf{\textit{Shortly after the theater shooting in 2012, he told ABC that the gunman was "diabolical" and would have found another way to carry out his massacre}}} \textcolor{red}{\textbf{if guns had not been available, a common argument from gun-control opponents}}.                                                                                                                                                                         &  Shortly after the theater shooting in 2012, he told ABC that \textcolor{example_blue}{\textbf{\textit{the gunman was "diabolical" and would have found another way to carry out his massacre}}} \textcolor{red}{\textbf{if guns had not been available}}, a common argument from gun-control opponents.                                                                                                                                                                        \\ \hline
    Now, \textcolor{red}{\textbf{if the priests in the Vatican had done their job in the first place, a quiet conversation, behind closed doors}} and \textcolor{example_blue}{\textbf{\textit{much of it would have been prevented}}}.                                                                                                                                                                                                                                             & Now, \textcolor{red}{\textbf{if the priests in the Vatican had done their job in the first place}}, \textcolor{example_blue}{\textbf{\textit{a quiet conversation, behind closed doors and much of it would have been prevented}}}.                                                                                                                                                                                                                                             \\ \hline
    The CPEC may have some advantages for Pakistan's economy -- for one, it has helped address the country's chronic power shortage -- but the costs are worrisome and \textcolor{red}{\textbf{unless they can be wished away with a wand}}, it will present significant issues in the future.                                                                                                                                          & The CPEC may have some advantages for Pakistan's economy -- for one, it has helped address the country's chronic power shortage -- but the costs are worrisome and \textcolor{red}{\textbf{unless they can be wished away with a wand}}, \textcolor{example_blue}{\textbf{\textit{it will present significant issues in the future}}}.                                                                                                                                          \\ \hline
    \end{tabular}
    }
    \caption{An example of bad predictions from LRM over Subtask 2 validation data. Antecedents are highlighted with red bold, consequents with blue bold italic.}
\end{table}

\end{document}